# Resume Evaluation through Latent Dirichlet Allocation and Natural Language Processing for Effective Candidate Selection

Vidhita Jagwani
Department of Computer Engineering
Sardar Patel Institute of Technology
Mumbai, India
vidhita.jagwani@spit.ac.in

Smit Meghani
Department of Computer Engineering
Sardar Patel Institute of Technology
Mumbai, India
smit.meghani@spit.ac.in

Krishna Pai
Department of Computer Engineering
Sardar Patel Institute of Technology
Mumbai, India
krishna.pai@spit.ac.in

Sudhir Dhage
Department of Computer Engineering
Sardar Patel Institute of Technology
Mumbai, India
sudhir_dhage@spit.ac.in

*Abstract-* *With the increasing number of job applicants, automated resume rating has become a necessity for recruiters. In this paper, we propose a method for resume rating using Latent Dirichlet Allocation (LDA) and entity detection with SpaCy. The proposed method first extracts relevant entities such as education, experience, and skills from the resume using SpaCy's Named Entity Recognition (NER). The LDA model then uses these entities to rate the resume by assigning topic probabilities to each entity. Furthermore, we conduct a detailed analysis of the entity detection using SpaCy's NER and report its evaluation metrics.*

*Using LDA, our proposed system breaks down resumes into latent topics and extracts meaningful semantic representations. With a vision to define our resume score to be more content-driven rather than a structure and keyword match driven, our model has achieved 77% accuracy with respect to only skills in consideration and an overall 82% accuracy with all attributes in consideration. (like college name, work experience, degree and skills)*

## I. INTRODUCTION

The constant news of mass layoffs from top technology companies has left many wondering the actual reason that spiked this rise of firing employees. Is there a bridge between the skills that a person has versus what the company actually wants? Well even though the reason for this bridge remains unclear, the obvious effect of this has been the deep emotional shock to the employees and a frenzy to find a new job, especially for those who live from paycheck to another.

The grave problem we identified in such cases of unforeseen layoffs, without any clarity, employees tend to psychologically blame themselves or their lack of skill for the same. Alongside this, while applying to other companies for jobs, there is no direct aid of letting a potential employee know how well suited they are for the job being offered. Regular rejections can be demotivating emotionally and many people often give up on their hope to find a good job in this rat race.

Since the emergence of job prospects in the technological domain, resume parsers have been a standard way of analyzing a person's eligibility for a job. We intend to display a mathematical score as a metric of evaluation of one's resume based on a set of attributes derived from the resume itself. In order to do so, we will use text mining using Latent Dirichlet Allocation (LDA) and entity detection with SpaCy. The score rating ranges from 0 to 10 and is used to provide a standardized and intuitive way of communicating the resume rating to the recruiters and users.

## II. OUTLINE OF THE PAPER

The chronology of the paper is as follows:-

Section III follows an in-depth timeline analysis of how resumes were initially reviewed and the reason for technological advancement in this process. It also explores the baseline model on which most resume scores are created and ways to modify and optimize it.

Section IV focuses on our proposed system for a resume scorer. It involves a high-level resume scorer workflow diagram, from the input by the user to the generated score output. Furthermore, it dives into the working of SpaCy's NER model for text-extraction from the resumes. From the extracted text, we have explained the logic of working of Latent Dirichlet Algorithm (LDA) based on Bayesian inference techniques. Finally, we have mentioned the mathematical logic to generate our score.

Section V demonstrates the output of our system including the word-topic distribution for each document using LDA, the resume score and the accuracy of our system.

Section VI summarizes the paper and highlights further scope of our project to optimize the output.

Section VII is the directory of references including research papers that aided in our research and resume score system development.

## III. LITERATURE SURVEY

The very early practice of resume viewing was a manual one, where a group of HR professionals scanned resumes to identify which ones best suited the job. Although this method was fruitful to identify diverse skills with human intervention, it wasn't the most efficient process as the number of resumes kept increasing. With over thousands of resumes to be scanned by big companies, it became increasingly time consuming to analyze each resume. Alongside this, manual resume rating is inherently subjective as it relies on individual interpretations and judgements of the reviewers. Different reviewers may have different criteria and preferences, leading to inconsistencies in the evaluation process. This subjectivity can introduce bias and impact the fairness of the selection process.

A more prevalent method of using feature extraction and keyword matching was used to automate the process of resume parsing. However, relying solely on keyword matching can be problematic as it may overlook important skills and experiences that are not explicitly mentioned in the resume. It may also result in false positives if candidates include keywords without actually possessing the required expertise.

The paper titled "CareerMapper: An automated resume evaluation tool,"[7] authored by V. Lai et al., presents CareerMapper as an automated resume evaluation tool that utilizes NLP techniques to analyze and extract information from resumes. The tool offers a practical solution for efficiently evaluating a large number of resumes, providing accuracy, efficiency, and time savings for recruiters. The research contributes to the field of automated recruitment processes and showcases the potential of NLP in improving hiring practices.

The paper titled "Feature Extraction and Analysis of Natural Language Processing for Deep Learning English Language,"[2] authored by D. Wang, J. Su, and H. Yu, presents a comprehensive analysis of feature extraction techniques in NLP for deep learning in the English language. The authors emphasize the significance of both traditional linguistic features and deep learning-based features and propose a combined approach. The experimental evaluation demonstrates the superiority of the proposed approach in improving the performance of deep learning models in various NLP tasks. The research contributes to advancing the field of NLP and provides insights for developing more effective NLP models.

During the research, we were also introduced to the involvement of unsupervised learning through LDA Algorithm in resume parsing. By using LDA in resume parsing, the algorithm can identify the underlying topics present in the resume content, enabling the extraction of specific information related to each topic. LDA works by considering the text within a resume document as a collection of words and identifying the latent topics that are likely to generate those words. It assumes that each resume is a mixture of multiple topics, and each word in the resume is associated with a particular topic. The goal of LDA is to uncover these latent topics and estimate the topic proportions for each resume.

The paper titled "Incorporating Lexical Priors into Topic Models", [10] authored by Jagadeesh Jagarlamudi, Hal Daume III ́ and Raghavendra Udupa, presents the concept of guided LDA. This is one of the futuristic scope of our model; it focuses on an effective way to guide topic models to learn topics of specific interest to a user. This is achieved by providing sets of seed words that a user believes are representative of the underlying topics in a corpus. The proposed model uses these seeds to improve both topic word distributions (by biasing topics to produce appropriate seed words) and to improve document-topic distributions (by biasing documents to select topics related to the seed words they contain).

## IV. PROPOSED RESUME RATING SYSTEM

The method proposed in this paper aims to leverage Latent Dirichlet Allocation (LDA) and Named Entity Recognition (NER) for rating and analysis of resumes. The combination of LDA for topic modeling and NER for entity extraction provides valuable insights into the themes, keywords, and important entities within the resumes, facilitating a more objective and informative assessment process. Fig-1 illustrates the key steps involved in our project

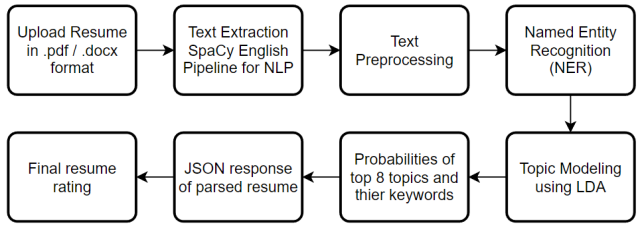

*Fig-1*: Resume Rating workflow block diagram

*A) Data collection and Text Extraction*

(i) Data Collection: Gathering a diverse dataset of resumes in PDF/DOCX format from various sources, representing different industries, job positions, and skill sets.

(ii) Text Extraction: Utilize Spacy, a powerful *NLP* library, to accurately extract the textual content from resumes while preserving the document structure. Apply text preprocessing techniques to clean the extracted text, removing special characters, punctuation, formatting, and stopwords.

(iii) Spacy's Language Model: Employ Spacy's pre-trained model, en_core_web_lg which provides essential *NLP* functionalities like tokenization, part-of-speech tagging, and named entity recognition *(NER)*. Leverage the model's word vectors to capture semantic and contextual information for better understanding of word meanings and relationships. The Fig-2 illustrates SpaCy's text preprocessing timeline.

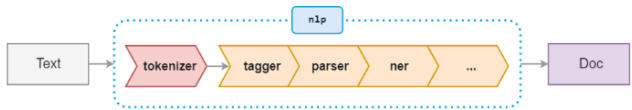

*Fig-2:* SpaCy text processing pipeline
(Source: https://spacy.io/usage/spacy-101)

*B) Topic Modeling with LDA and keyword extraction*

(i) LDA Model: The LDA model is applied to the training set after being trained on the preprocessed dataset. Each resume receives a topic based on the distribution of words inside it according to the LDA model. The top themes for each resume may be found by looking at the LDA model's output, along with their related keywords and the likelihoods that each topic will be present. This information gives important insights on the primary topics and subject matter of the resumes.

(ii) Named Entity Recognition (NER): The methodology uses named entity recognition techniques in addition to topic modeling to extract crucial entities from the resumes. NER algorithms recognise and categorize things, such as names, contact information, educational background, and skills listed in resumes. Key data can be retrieved and used for analysis and assessment by utilizing NER.

*C) Resume Parsing and Rating*

(i) Resume Parsing: The NER component of Spacy's en_core_web_lg model is trained and employed to identify and extract important entities from the resume text. The parsing accurately identifies and extracts the desired resume sections, such as contact details, location, education, work experience, skills.

(ii) Resume Rating: We employ matching strategies to match keywords to the document to fulfill (1) in domain knowledge and to match words inside the document to fulfill (2) after utilizing the LDA model to identify the keywords and their probabilities.
(1) Keyword Match (KM): Look for match in keywords
(2) Within Match (WM): Look for match within words matched by keywords (Conciseness of experience)

$$finalScore = KM * WM \quad (1)$$
$$diff = (finalScore - mean)/sd \quad (2)$$
$$rating = min(10, max(0, 5 + diff)) \quad (3)$$

The Fig-3 illustrates a JSON output of a sample resume that we gave input to our system. As we can identify clearly, via SpaCy's text extraction, we have been able to define attributes. The attributes are of 2 types. First, the ones that amount to personal information, like- name, email, number, city. And second, those which contribute to the score like work experience, education, work and education duration and skill set. Finally we notice a score output at the end, which is a metric out of 10.

```
{
    "name": "John Doe (from filename)",
    "email": "cmcturland@email.com",
    "number": "(123) 456-7890",
    "city": "New York",
    "work_exp:": [
        "Constructed the logic for a streamlined ad-serving platform that",
        "Tested software for bugs and operating speed, fixing bugs and",
        "Iterated platform for college admissions, collaborating with a group",
        "Software Engineer",
        "MarketSmart",
        "April 2012 - January 2015 / Washington, DC",
        "Built RESTful APIs that served data to the JavaScript front-end",
        "Built internal tool using NodeJS and Pupeteer.js to automate QA and",
        "Reviewed code and conducted testing for 3 additional features on",
        "Software Engineer Intern",
        "Marketing Science Company",
        "April 2011 - March 2012 / Pittsburgh, PA",
        "Partnered with a developer to implement RESTful APIs in Django,",
        "Using Selenium I built out a unit testing infrastructure for a client"
    ],
    "education": [
        "B.S. Computer Science",
        "University of Pittsburgh",
        "September 2008 - April 2012",
        "Pittsburgh, PA"
    ],
    "work_duration": "3 years 8 months",
    "education_duration": "3 years 7 months",
    "skills": "Nodejs, Open, Analytics, Reporting, Apis, Javascript, Python, Marketing, Aws, Html5, Unix, Django, Sci, Sql, Admissions, Postgresql, Selenium, Mysql, Git, Testing, Nosql, Css, Reactjs, Jquery",
    "rating": 8.13
}
```

*Fig-3:* Parsed sample resume along with rating

*Latent Dirichlet Allocation (LDA):* LDA is a generative probabilistic model for topic modeling. It assumes that each document in a collection is a mixture of different topics, and each topic is characterized by a distribution of words. LDA aims to uncover these latent topics by estimating the underlying topic distributions within the documents.

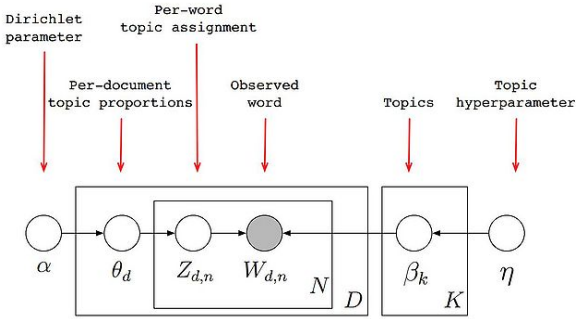

*Fig-4:* Latent Dirichlet Allocation: Generative Model

$draw\ each\ topic\ \beta_i \sim Dir(\eta)\ for\ i \in \{1...K\}$
α: Dirichlet parameter to draw
$each\ document\ \theta_d \sim Dir(\alpha)\ and\ W_{d,n} \sim Mult(\beta z_{d,n})$
Total of K topics, D documents, N words

To estimate these distributions, LDA employs Bayesian inference techniques, such as variational inference or Gibbs sampling. These methods iteratively update the distributions based on the observed words and prior assumptions.

By learning the topic distributions, LDA enables the identification of latent topics within documents, which can be used for tasks like document clustering, topic modeling, or in the context of resume rating, identifying relevant themes in resumes based upon attributes that affect the score.

## V. RESULT AND ANALYSIS

**Performance of Entity Detection using SPaCy's NER:**
The model is tested on 20 resumes, and the projected resume summaries are saved separately for each resume as .txt files. We determine the accuracy score, precision, recall, and f-score for each entity that the model is able to identify for each resume on which it is tested. For each object, the values of these metrics are added up and averaged to produce an overall score that is used to assess the model on the test set of 20 resumes. The findings of the entity-wise evaluation are shown below in Table-1. We can notice that the accuracy to be 82% due to the effect of all entities and about 77% by just the skills.

| Entity | precision | recall | F1-score |
|---|---|---|---|
| College Name | 0.74 | 0.78 | 0.759 |
| Degree | 0.91 | 0.88 | 0.895 |
| Email | 1.00 | 0.86 | 0.925 |
| location | 0.65 | 0.53 | 0.584 |
| Name | 0.95 | 0.91 | 0.930 |
| Skills | 0.88 | 0.67 | 0.761 |

**Table-1:** Automatic Summarization of Resumes with SpaCy's Named Entity Recognition (NER)

Setting a dropout rate—a rate at which specific features and representations are randomly "dropped"—is another method for enhancing learning outcomes. As a result, the model has a tougher time remembering the training data. A 0.25 dropout, for instance, indicates that each feature or internal representation has a 1/4 chance of being lost. We train the model for 10 epochs while maintaining a 0.2 dropout rate.

**LDA Model Performance and Comparison with Baseline:**
We employ Latent Dirichlet Allocation (LDA), a generative model that presupposes that every document d is a conglomerate of a few subjects z and that each word in the text is attributed to one of the topics[8]. To put it another way, it is assumed that d is produced by keywords drawn from a variety of themes Z.

It computes, P(Z|D=d), the probability that for each z in Z, d is generated, e.g.:
CV of Ben Lee: [z_1: 0.2, z_2: 0.3, z_3: 0.1, z_4: 0.4]

P(K|Z), the probability that a word k of K is picked from topic z of Z, e.g. as shown in Fig-5.

```
z_1: [computer: 0.3, engineering: 0.2, leadership: 0.1, software: 0.05, ...] == 1
z_2: [finance: 0.33, quantitative: 0.21, leadership: 0.1, asset: 0.08, ...] == 1
z_3: [management: 0.12, general: 0.1, report: 0.07, lead: 0.05, ...] == 1
z_4: [human: 0.22, resource: 0.21, recruitment: 0.1, award: 0.08, ...] == 1
```

*Fig-5:* Parsed sample resume along with rating

We then model this as a Bayesian Network and then find the keywords. All we need to do is select n to specify how many keywords we want for each document d. For the sample resume we set n=30 and the nargmax P(K|D=d) is calculated:

```
{
  "cost": "0.027297724",
  "reduce": "0.02560255",
  "deliver": "0.024087662",
  "delivery": "0.023979329",
  "lead": "0.021393068",
  "migrate": "0.018958531",
  "business": "0.018151239",
  "day": "0.015336814",
  "topic_score": "0.046086058",
  "topic": "26"
}
{
  "app": "0.08347968",
  "backend": "0.081356",
  "engineer": "0.06869899",
  "software engineer": "0.0600872",
  "software": "0.04070151",
  "linkedin": "0.03966418",
  "android": "0.03692913",
  "stack": "0.026822906",
  "topic_score": "0.045362577",
  "topic": "13"
}
{
  "source": "0.074811265",
  "concept": "0.0549722",
  "bachelor": "0.05470456",
  "process": "0.046730053",
  "write": "0.032443535",
  "upload": "0.029692635",
  "software": "0.025725719",
  "test": "0.02571591",
  "topic_score": "0.03450413",
  "topic": "15"
}
{
  "website": "0.086609416",
  "web": "0.05175313",
  "portal": "0.047178593",
  "mysql": "0.046417154",
  "php": "0.040003475",
  "developer": "0.03267589",
  "javascript": "0.0295029",
  "bootstrap": "0.02804043",
  "topic_score": "0.03246912",
  "topic": "8"
}
```

*Fig-6:* The word topic distribution for top four topics

**Rating Distribution and Analysis:**
With reference to equations (1), (2) and (3)
For instance, rating is equal to **rating** = 5 + (0.4-0.3)/0.05 = 7 if **final_score** is **0.4**, **mean** is **0.3**, and **standard deviation** is **0.05**. Since we don't actually expect scores to deviate more than 5 standard deviations from the mean, we have a min and max to limit the scores between 0 and 10. This makes sense. Because the scores are fairly regularly distributed, we may apply this formula. Low intra-model variation also contributes to the validity of the score's application.

## VI. CONCLUSION AND FUTURE SCOPE

This research paper has presented a novel approach for automating the resume rating process by leveraging Latent Dirichlet Allocation (LDA) and SpaCy's Named Entity Recognition (NER). By integrating NER, the identification of key entities such as education, experience, and skills from resumes has been enhanced, leading to more accurate ratings. The LDA model has effectively assigned topic probabilities to these entities, enabling the generation of a final score for resume evaluation. The experimental results demonstrate the effectiveness of the proposed method, surpassing the performance of the baseline model with an accuracy of 82%.

There are several areas for further enhancement and exploration.
(i) Expanding the dataset size and diversity across multiple career domains would provide a more comprehensive and versatile evaluation platform.
(ii) Investigating different variations of LDA models such as *guided LDA* or exploring alternative machine learning techniques like *BERT* could potentially improve the accuracy and robustness of resume rating.
(iii) Incorporating user feedback and iterative model refinement can contribute to continuous improvement of the system. Overall, this research lays a foundation for further advancements in automated resume rating and highlights avenues for future research and development.